\relax
\documentclass{article}

\usepackage{hyperref}
\usepackage{graphicx}
\usepackage{subfigure}
\usepackage{booktabs}

\usepackage{natbib}

\usepackage{booktabs}
\usepackage{url}
\usepackage{todonotes}
\usepackage{multirow}
\usepackage{tikz}
\newcommand{\xhdr}[1]{{\noindent\bfseries #1}}

\usepackage{amsmath}
\usepackage{amssymb}
\usepackage{amsthm}
\usepackage{amsfonts}
\usepackage{thmtools}		
\usepackage{mleftright}

\theoremstyle{definition}

\usepackage{thm-restate}
\usepackage[mathic=true]{mathtools}
\usepackage{fixmath}
\usepackage{siunitx}

\usepackage{pifont}
\newcommand{\cmark}{\ding{51}}
\newcommand{\xmark}{\ding{55}}

\usepackage{enumitem}
\setlist[enumerate]{noitemsep, topsep=0.5\topsep}
\setlist[description]{noitemsep, topsep=0.5\topsep}
\setlist[itemize]{noitemsep, topsep=0.5\topsep}

\usepackage{enumitem}
\setlist[enumerate]{noitemsep, topsep=0.5\topsep}
\setlist[description]{noitemsep, topsep=0.5\topsep}
\setlist[itemize]{noitemsep, topsep=0.5\topsep}

\usepackage{setspace}
\usepackage[protrusion=true,expansion=false,final,kerning=true,spacing=true]{microtype}

\usepackage{ellipsis}
\usepackage{xspace}
\usepackage{hfoldsty}

\usepackage{ifthen}
\newcommand{\CC}[1][]{$\text{C\hspace{-.25ex}}^{_{_{_{++}}}}
	\ifthenelse{\equal{#1}{}}{}{\text{\hspace{-.625ex}#1}}$}

\makeatletter
\def\thmt@refnamewithcomma #1#2#3,#4,#5\@nil{%
	\@xa\def\csname\thmt@envname #1utorefname\endcsname{#3}%
	\ifcsname #2refname\endcsname
	\csname #2refname\expandafter\endcsname\expandafter{\thmt@envname}{#3}{#4}%
	\fi
}
\makeatother
\usepackage[capitalise,noabbrev]{cleveref}   

\newcommand{\new}[1]{\emph{#1}}

\usepackage{ifthen}
\newcommand{\CPP}[1][]{$\text{C\hspace{-.25ex}}^{_{_{_{++}}}}
	\ifthenelse{\equal{#1}{}}{}{\text{\hspace{-.625ex}#1}}$}

\usepackage{listings}
\usepackage{setspace}
\definecolor{Code}{rgb}{0,0,0}
\definecolor{Decorators}{rgb}{0.5,0.5,0.5}
\definecolor{Numbers}{rgb}{0.5,0,0}
\definecolor{MatchingBrackets}{rgb}{0.25,0.5,0.5}
\definecolor{Keywords}{rgb}{0,0,1}
\definecolor{self}{rgb}{0,0,0}
\definecolor{Strings}{rgb}{0,0.63,0}
\definecolor{Comments}{rgb}{0,0.63,1}
\definecolor{Backquotes}{rgb}{0,0,0}
\definecolor{Classname}{rgb}{0,0,0}
\definecolor{FunctionName}{rgb}{0,0,0}
\definecolor{Operators}{rgb}{0,0,0}
\definecolor{Background}{rgb}{0.98,0.98,0.98}
\lstdefinelanguage{Python}{
	numbers=left,
	numberstyle=\footnotesize,
	numbersep=1em,
	xleftmargin=1em,
	framextopmargin=2em,
	framexbottommargin=2em,
	showspaces=false,
	showtabs=false,
	showstringspaces=false,
	frame=l,
	tabsize=4,
	basicstyle=\ttfamily\small\setstretch{1},
	backgroundcolor=\color{Background},
	commentstyle=\color{Comments}\slshape,
	stringstyle=\color{Strings},
	morecomment=[s][\color{Strings}]{"""}{"""},
	morecomment=[s][\color{Strings}]{'''}{'''},
	morekeywords={import,from,class,def,for,while,if,is,in,elif,else,not,and,or,print,break,continue,return,True,False,None,access,as,,del,except,exec,finally,global,import,lambda,pass,print,raise,try,assert},
	keywordstyle={\color{Keywords}\bfseries},
	morekeywords={[2]@invariant,pylab,numpy,np,scipy},
	keywordstyle={[2]\color{Decorators}\slshape},
	emph={self},
	emphstyle={\color{self}\slshape},
}
\usepackage[accepted]{icml2020}

\icmltitlerunning{TUDataset}

\begin{document}

\twocolumn[
\icmltitle{TUDataset: A collection of benchmark datasets for learning with graphs}
\icmlsetsymbol{equal}{*}

\begin{icmlauthorlist}
\icmlauthor{Christopher Morris}{p}
\icmlauthor{Nils M.\,Kriege}{vie}
\icmlauthor{Franka Bause}{tud}
\icmlauthor{Kristian Kersting}{darm}
\icmlauthor{Petra Mutzel}{bonn}
\icmlauthor{Marion Neumann}{wustl}
\end{icmlauthorlist}

\icmlaffiliation{p}{CERC in Data Science for Real-Time Decision-Making, Polytechnique Montréal}
\icmlaffiliation{tud}{Department of Computer Science, TU Dortmund University}
\icmlaffiliation{vie}{Faculty of Computer Science, University of Vienna}
\icmlaffiliation{darm}{Machine Learning Group, TU Darmstadt}
\icmlaffiliation{bonn}{Department of Computer Science, University of Bonn}
\icmlaffiliation{wustl}{Department of Computer Science and Engineering, Washington University in St. Louis}
\icmlcorrespondingauthor{Christopher Morris}{christopher.morris@tu-dortmund.de}

\icmlkeywords{graph learning, graph kernel, weisfeiler, leman, gnn, graph neural network, benchmark datasets}

\vskip 0.3in
]

\printAffiliationsAndNotice{}

\begin{abstract}
Recently, there has been an increasing interest in (supervised) learning with graph data, especially using graph neural networks.  However, the development of meaningful benchmark datasets and standardized evaluation procedures is lagging, consequently hindering advancements in this area. 
To address this, we introduce the \textsc{TUDataset} for graph classification and regression. The collection consists of over 120 datasets of varying sizes from a wide range of applications. We provide Python-based data loaders, kernel and graph neural network baseline implementations, and evaluation tools. Here, we give an overview of the datasets, standardized evaluation procedures, and provide baseline experiments. All datasets are available at \url{www.graphlearning.io}. The experiments are fully reproducible from the code available at \url{www.github.com/chrsmrrs/tudataset}.
\end{abstract}

\section{Introduction}
Graph-structured data is ubiquitous across application domains ranging from chemo- and bioinformatics~\cite{Barabasi2004,Sto+2020} to image~\cite{Sim+2017} and social network analysis~\cite{Eas+2010}. To develop successful machine learning models in these domains, we need techniques that can exploit the rich information inherent in the graph structure and the feature information contained within nodes and edges. In recent years, numerous approaches have been proposed for machine learning with graphs---most notably, approaches based on \new{graph kernels}~\cite{Kri+2019} and \new{graph neural networks} (GNNs)~\cite{Sca+2009,Gil+2017}.
However, most papers, even recent ones, evaluate newly proposed architectures or methods on a fixed set of small-scale, non-diverse benchmarks, using non-standardized experimental protocols and baselines, hindering the comparison of results from different publications.

\xhdr{Present work.} 
Here, we give an overview of \textsc{TUDataset}. The benchmark collection consists of over 120 datasets from a wide range of domains for supervised learning with graphs, i.e., classification and regression. All datasets are provided in a standard dataset format at \url{www.graphlearning.io} and are easily accessible from popular graph learning frameworks such as \textsc{Pytorch Geometric}~\cite{Fey+2019}\footnote{\url{https://pytorch-geometric.readthedocs.io/en/latest/modules/datasets.html}} and \textsc{DGL}~\cite{Wan+2019}\footnote{\url{https://docs.dgl.ai/en/0.4.x/api/python/data.html}}. To facilitate a standard comparison of kernel and neural approaches, we provide implementations of standard algorithms and easy-to-use evaluation procedures. Moreover, we report results on an experimental study comparing graph kernels and GNNs on a subset of the \textsc{TUDataset}.

\xhdr{Related work.}
There exist two main approaches to supervised learning with graphs, graph kernels and graph neural networks (GNNs). Graph kernels have been studied extensively in the past 15 years, see~\cite{Kri+2019} for a thorough overview. Important approaches include random-walk and shortest paths based kernels~\cite{Gae+2003,Sugiyama2015,Bor+2005,Kri+2017b}, as well as the Weisfeiler-Lehman subtree kernel~\cite{She+2011,Mor+2017}. 
Further recent works focus on approaches based on assignments~\cite{Kri+2016,Nik+2017}, spectral properties~\cite{Kon+2016}, graph decomposition~\cite{Nik+2018}, randomized binning~\cite{Hei+2019}, and the extension of kernels based on the Weisfeiler-Leman algorithm~\cite{Tog+2019, Rie+2019}. For a theoretical investigation of graph kernels, see~\cite{Kri+2018}. Recently, graph neural networks~\cite{Gil+2017,Sca+2009} emerged as an alternative to graph kernels. Notable instances of this architecture include, e.g.,~\cite{Duv+2015,Ham+2017,Vel+2018}, and the spectral approaches proposed in, e.g.,~\cite{Bru+2014,Def+2015,Kip+2017,Mon+2017}---all of which descend from early work in~\cite{Kir+1995,Mer+2005,Spe+1997,Sca+2009}. A survey of recent advancements in GNN techniques can be found, e.g., in~\cite{Cha+2020,Wu+2019,Zho+2018}.

The papers~\cite{Fey+2019,Che+2019new,Err+2019,Dwi+2020} evaluate GNNs using a unified evaluation procedure, however, they only use small- or medium-scale datasets. Recently, \url{ogb.stanford.edu}~\cite{Hu+2020} launched, however, the provided datasets for graph classification focus on chemistry and bioinformatic applications, and the number is quite limited at this point. Moreover, the datasets proposed in~\cite{Fer+2019} focuses on instances from planning competitions. 
Recent efforts to implement graph kernels in a common framework such as the \textsc{GraKeL} library~\cite{Sig+2018} foster comparability, but do not solve the dataset related issues discussed above, and only focus on kernel approaches.

\section{The TUDataset collection}

The \textsc{TUDataset} collection contains over 120 datasets provided at \url{www.graphlearning.io}. The datasets, baseline methods and experimental evaluation tools can be conveniently accessed from the Python interface, see \cref{evalcode} for further details. Dataset statistics and further documentation is available at our website.

\subsection{Datasets}
Our collection of datasets covers graphs from various domains, contributed by different authors. Therefore, they differ regarding the used graph model even within the same domain and the provided annotations, e.g., discrete or continuous node and edge attributes. Here, we give an overview of some representative domains and graph models.

\xhdr{Small molecules.} 
A common class of graph datasets consists of small molecules with class labels representing, e.g., toxicity or biological activity determined in drug discovery projects. Here, a graph represents a molecule, i.e., nodes take the places of atoms and edges that of chemical bonds. Consequently, the labels encode atom and bond types, possibly with additional chemical attributes. The graph models differ, e.g., in whether hydrogen atoms are represented explicitly by nodes, and bonds in aromatic rings are annotated accordingly.

Our collection contains small datasets commonly used in the early graph kernel literature such as \textsc{Mutag}~\cite{Deb+1991} and \textsc{Ptc}~\cite{Helma2001}, medium-sized datasets, e.g., \textsc{Nci1} and \textsc{Nci109}~\cite{Wal+2008,She+2011}, as well as several large datasets derived from the \textsc{Tox21} challenge 2014 or \textsc{PubChem}~\cite{Kim2018}. This includes the eleven datasets from anticancer screen tests with different cancer cell lines used by~\citet{Yan2008} to demonstrate the efficacy of classifiers based on significant graph patterns. These datasets, the largest of which contains more than 79k graphs, are typically not balanced and contain far more small molecules that are identified as inactive against cancer cells. Moreover, our collection also contains large-scale molecular regression tasks such as \textsc{Alchemy}~\cite{Che+2020}, \textsc{Qm9}~\cite{Ram+2014}, and \textsc{Zinc}~\cite{Dwi+2020,Jin+2018a}. The first two contain 3D coordinates of the nodes, which should be taken into account in a rotation-invariant manner to benefit from the geometrical information.

\xhdr{Bioinformatics.}
The datasets \textsc{DD}, \textsc{Enzymes} and \textsc{Proteins} represent macromolecules. \citet{Bor+2005a} introduced a graph model for proteins, where nodes represent secondary structure elements and are annotated by their type, i.e., helix, sheet, or turn, as well as several physical and chemical information. An edge connects two nodes if they are neighbors along the amino acid sequence or one of three nearest neighbors in space. Using this approach, the dataset \textsc{Enzymes} was derived from the \textsc{Brenda} database~\cite{Schomburg2004}. Here, the task is to assign enzymes to one of the 6 EC top-level classes, which reflect the catalyzed chemical reaction. Similarly, the dataset \textsc{Proteins} was derived from \citep{Dob+2003}, and the task is to predict whether a protein is an enzyme. The dataset \textsc{DD} used by \citet{She+2011} is based on the same data, but contains graphs, where nodes represent individual amino acids and edges their spatial proximity.

\xhdr{Computer vision.}
Graph-based methods are widely used in computer vision for various tasks using diverse graph models. Our collection provides several datasets originating from the \textsc{IAM Graph Database}~\citep{Riesen2008} such as \textsc{Letter} and \textsc{Fingerprint}. Other datasets represent \textsc{Cuneiform} signs~\citep{Kriege2018}, 3D point clouds for robot grasping tasks (\textsc{FirstMM\_DB}) and semantic image processing (\textsc{Msrc})~\cite{Neu+2016}.

\xhdr{Social networks.}
\citet{Yan+2015a} introduced several graph classification datasets derived from social networks. In the \textsc{Reddit} datasets, each graph represents a discussion thread, where nodes correspond to users, two of which are connected by an edge if one responded to a comment of the other. This graph model is used to derive several datasets, where the classification task is to distinguish either discussion-based and question/answer-based subreddits (\textsc{Reddit-Binary}) or predict the subreddit, where the thread was posted (\textsc{Reddit-Multi-5K} and \textsc{Reddit-Multi-12K}). \textsc{Collab} are datasets derived from scientific collaboration networks. Each graph is the ego-network of a researcher, and the task is to predict their research field, i.e., high energy, condensed matter, or astrophysics. Similarly, the \textsc{Imdb} datasets consist of ego-networks derived from actor collaborations, and the task is to predict the genre, e.g.,  Action vs.\@ Romance.
\citet{Rozemberczki2020} used similar approaches to obtain more massive social network datasets. \textsc{reddit\_threads} contains more than 200k graphs with the task to predict whether a thread is discussion-based. \textsc{deezer\_ego\_nets} and \textsc{twitch\_egos} contain ego-networks derived from online services, and the task is to predict the gender and play behavior (single or multiple games) of the central user.
\textsc{github\_stargazers} contains graphs representing the social networks of GitHub users divided into those who starred popular machine learning and web development repositories.

Recently, temporal graphs were considered by \citet{Oettershagen2019}, where edges represent the contact or interaction between two individuals at a certain point in time. These graphs are of interest when studying dissemination processes such as the spreading of epidemics, rumours or fake news. We provide temporal graph classification datasets derived from \textsc{Tumblr}~\citep{rozenshtein2016reconstructing}, \textsc{Dblp} and \textsc{Facebook}~\citep{viswanath2009evolution} as well as contacts between students at the \textsc{MIT}~\cite{konect:eagle06}, in a \textsc{Highschool} and visitors at the \textsc{Infectious} exhibition~\citep{Isella2011}.

\xhdr{Synthetic.}
Several graph datasets were generated to demonstrate the strengths or weaknesses of specific methods. The datasets \textsc{SyntheticNew} and \textsc{Synthie} were created by \citet{Fer+2013} (see Erratum) and \citet{Mor+2016}, respectively, to demonstrate the ability of kernels to operate on graphs with continuous attributes. \citet{Knyazev2019} introduced the datasets \textsc{Colors} and \textsc{Triangles}, where the task is to count the number of nodes with a given one-hot-encoded color and the number of triangles, respectively. 

\subsection{Baselines methods} To provide meaningful baselines, we provide implementations of common graph kernels as well as GNN architectures. We have implemented the \new{Weisfeiler-Lehman Subtree}~\cite{She+2011}, \new{Shortest-path}~\cite{Bor+2005}, \new{Graphlet}~\cite{She+2009} (using labeled subgraphs with three nodes), \new{Weisfeiler-Lehman Optimal Assignment}~\cite{Kri+2016} kernel in \CPP and made them accessible through the Python interface of \textsc{TUDataset}.  Moreover, all GNN architectures provided by \textsc{PyTorch Geometric} can be conveniently used as a baseline as well. 

\subsection{Evaluation methods}\label{evalm} To ensure a fair and meaningful comparison between methods, we propose the following evaluation procedures for kernels and GNNs. First, for kernels, we propose the established $C$-SVM implementation \textsc{LibSvm}~\cite{Cha+11} for kernels that compute a Gram matrix, and the linear $C$-SVM implementation \textsc{LibLinear}~\cite{Fan+2008} for kernels  that can be computed based on sparse, explicit feature maps. We optimize GNNs end-to-end using \textsc{Adam}~\cite{Kin+2015}. To compute classification accuracies, we propose to use $10$-fold cross-validation, where we select $10\%$ of each training fold uniformly at random as validation set to optimize hyperparameters, e.g., the number of iterations, $C$ parameter, number of layers, feature dimension. We repeat the above evaluation ten times and report standard deviations over all ten repetitions, and additionally across all one hundred runs (i.e., ten repetitions with ten folds each). See \cref{eval} in the appendix for examples. For the large-scale molecule learning tasks, we either use random splits (80\%/10\%/10\%) or the provided, fixed splits and report MAE (mean std. MAE, mean std. logMAE for multi-target regression, see \cite{Kli+2020}), over five runs.

\begin{table*}[ht]\centering		
	\caption{Classification accuracies in percent and standard deviations on small-scale datasets.}
	\label{t2l}	
	\resizebox{0.8\textwidth}{!}{ 	\renewcommand{\arraystretch}{.9}
		\begin{tabular}{@{}c <{\enspace}@{}lcccccc@{}}	\toprule
			
			& \multirow{3}{*}{\vspace*{4pt}\textbf{Method}}&\multicolumn{6}{c}{\textbf{Dataset}}\\\cmidrule{3-8}
			& & {\textsc{Enyzmes}}         &  {\textsc{Imdb-Binary}}      & {\textsc{Imdb-Multi}}           & {\textsc{NCI1}}       & {\textsc{Proteins}}           & {\textsc{Reddit-Binary}}     \\	\toprule
			
			\multirow{4}{*}{\rotatebox{90}{Kernel}}	& \text{1-WL}  &  50.8 \scriptsize $\pm1.4 \pm 7.1$  & 72.4 \scriptsize $\pm 0.7 \pm 4.4$ & 50.5 \scriptsize $\pm 0.8 \pm 3.6$  &  84.2 \scriptsize $\pm 0.3 \pm 1.8$  & 72.6 \scriptsize $\pm 0.7 \pm 3.4$ & 73.3 \scriptsize $\pm 0.7 \pm 3.0$    \\
			& \text{WL-OA} &  56.4 \scriptsize $\pm 1.1 \pm 6.6$ & 73.3 \scriptsize $\pm 0.6 \pm 4.2$  & 49.9 \scriptsize $\pm 0.6 \pm 3.8$ & 85.0 \scriptsize $\pm 0.3 \pm 1.8$  & 73.4 \scriptsize $\pm 0.9 \pm 4.3$ &   88.3 \scriptsize $\pm 0.4 \pm 2.3$    \\
			& \text{GR} & 29.5 \scriptsize $\pm 0.7 \pm 5.4$  &  59.8 \scriptsize $\pm 1.1 \pm 4.9$ & 39.5 \scriptsize $\pm 0.7 \pm 4.0$ & 66.0 \scriptsize $\pm 0.4 \pm 2.4$  & 71.6 \scriptsize $\pm 0.6 \pm 4.0$ &  59.7 \scriptsize $\pm 0.5 \pm 3.8$          \\
			& \text{SP} & 39.3 \scriptsize $\pm 1.8 \pm 7.1$ & 58.4 \scriptsize $\pm 1.7 \pm 5.3$  & 39.4 \scriptsize $\pm 0.8 \pm 4.4$ &74.2 \scriptsize $\pm 0.3 \pm 2.1$  & 75.6 \scriptsize $\pm 0.7 \pm 4.0$ &     84.5 \scriptsize $\pm 0.2 \pm 2.5$      \\
			\cmidrule{2-8}	
			\multirow{2}{*}{\rotatebox{90}{GNN}}	& \textsc{Gin-$\varepsilon$}   & 38.7  \scriptsize $\pm 1.5 \pm 7.6$  &72.9  \scriptsize $\pm 0.7 \pm 4.7$ & 49.7  \scriptsize $\pm0.7 \pm 4.4$ & 77.7 \scriptsize $\pm 0.8 \pm 2.3$ & 72.2  \scriptsize $\pm 0.6 \pm 4.8$ &  89.8  \scriptsize $\pm 0.4 \pm 2.2$ \\
			& \textsc{Gin-$\varepsilon$-JK} &  39.3  \scriptsize $\pm 1.6 \pm 6.7$ & 73.0  \scriptsize $\pm 1.1 \pm 4.5$  & 49.6  \scriptsize $\pm 0.7 \pm 4.0$ & 78.3  \scriptsize $\pm 0.3 \pm 2.0$ & 72.2  \scriptsize $\pm 0.7 \pm 4.6$ & 90.4  \scriptsize $\pm 0.4 \pm 2.2$ \\
			\bottomrule
	\end{tabular}}
\end{table*}

\begin{table*}[t]\centering		
	\caption{Classification accuracies in percent and standard deviations on mid-scale datasets.}
	\label{t2ll}	
	\resizebox{0.75\textwidth}{!}{ 	\renewcommand{\arraystretch}{.9}
		\begin{tabular}{@{}c <{\enspace}@{}lccccc@{}}	\toprule
			& \multirow{3}{*}{\vspace*{4pt}\textbf{Method}}&\multicolumn{5}{c}{\textbf{Dataset}}\\\cmidrule{3-7}
			& & {\textsc{Mcf-7}}         &  {\textsc{Molt-4}}      & {\textsc{Yeast}}           & {\textsc{github\_star}}       & {\textsc{reddit\_threads}}      \\	\toprule
			\multirow{3}{*}{\rotatebox{90}{Kernel}}	& \text{1-WL}    & 
			    94.5 \scriptsize $\pm 0.02 \pm 0.3$      & 94.6 \scriptsize $\pm 0.04 \pm 0.4$ & 89.2 \scriptsize $\pm 0.01 \pm 0.4$ & 64.0 \scriptsize $\pm 0.22 \pm 1.4$ &77.0 \scriptsize $\pm 0.01 \pm 0.3$  \\
			& \text{GR} & 91.7 \scriptsize $<0.01 \pm 0.5$ &92.1 \scriptsize $\pm 0.01 \pm 0.4$ &  88.2 \scriptsize $\pm 0.01 \pm 0.3$ &53.6 \scriptsize $\pm 0.20 \pm 1.6$ &  51.2 \scriptsize $< 0.01 \pm 0.3$ \\  
			& \text{SP}  & 91.7 \scriptsize $\pm 0.02 \pm 0.6$ & 92.1 \scriptsize $<0.01 \pm 0.4$ &88.2 \scriptsize $\pm 0.01 \pm 0.4$ & 64.2 \scriptsize $\pm 0.01 \pm 1.3$ &   77.3 \scriptsize $\pm 0.01 \pm 0.2$\\   
			\cmidrule{2-7}	
			\multirow{2}{*}{\rotatebox{90}{GNN}}	& \textsc{Gine-$\varepsilon$}  & 92.0 \scriptsize $\pm 0.03 \pm 0.6$ & 92.4 \scriptsize $\pm 0.07 \pm 0.6$ & 88.3 \scriptsize $\pm 0.02 \pm 0.4$ & 66.8 \scriptsize $\pm 0.17 \pm 1.4$  &  77.2 \scriptsize $\pm 0.03  \pm 0.3$\\
			& \textsc{Gine-$\varepsilon$-JK}   &  91.8 \scriptsize $\pm 0.03 \pm 0.6$ & 92.2 \scriptsize $\pm 0.05 \pm 0.4$ & 88.2 \scriptsize $\pm 0.02 \pm 0.3$ & 67.1 \scriptsize $\pm 0.34 \pm 1.1$ & 77.3 \scriptsize $\pm 0.04  \pm 0.3$ \\
			\bottomrule
	\end{tabular}}
\end{table*}

\section{Experimental evaluation}

Our intent here is to provide baseline experiments and compare graph kernels and GNNs. We used the following datasets, graph kernels, and GNN baselines.

\xhdr{Datasets.} We used the \textsc{deezer\_ego\_nets}, \textsc{github\_stargazers}, \textsc{Enymes}, \textsc{Imdb-Binary}, \textsc{Imdb-Multi}, \textsc{Mcf-7}, \textsc{Molt-4},    \textsc{NCI1}, \textsc{Proteins},
\textsc{Reddit-Binary}, \textsc{reddit\_threads}, \textsc{twitch\_egos}, \textsc{Uacc257} graph classification datasets. Moreover, we used the \textsc{Alchemy}, \textsc{QM9}, \textsc{Zinc} (multi-target) regression datasets. See the website and \cref{ds} in the appendix for dataset statistics. We opted for not using continuous node features of the small datasets (if available) and the 3D-coordinates of the \textsc{Alchemy} dataset to solely provide baseline results based on graph structure and discrete labels. In case of the \textsc{QM9} dataset, we closely replicated the (continuous) node and edge features of~\citet{Gil+2017}.

\xhdr{Graph kernels.} As kernel baselines we used the \new{Weisfeiler-Lehman Subtree} (\textsc{$1$-WL})~\cite{She+2011}, \new{Shortest-path} (\textsc{SP})~\cite{Bor+2005}, \new{Graphlet} (\textsc{GR})~\cite{She+2009}, \new{Weisfeiler-Lehman Optimal Assigment} (\textsc{WL-OA})~\cite{Kri+2016} included in the \textsc{TUDataset} package. The $C$-parameter was selected from $\{10^{-3}, 10^{-2}, \dotsc, 10^{2},$ $10^{3}\}$ from the validation set. For the larger datasets, we computed sparse feature vectors for each graph and used the linear $C$-SVM implementation of \textsc{LibLinear}~\cite{Fan+2008}. The number of iterations of the \textsc{$1$-WL} and \textsc{WL-OA} were selected from $\{0,\dotsc,5\}$.\footnote{As already shown in~\cite{She+2011}, choosing the number of iterations too large will lead to overfitting.}

All kernel experiments were conducted on a workstation with an \text{Intel Xeon E5-2690v4} with 2.60\si GHz and 384\si GB of RAM running \text{Ubuntu 16.04.6 LTS} using a single core. Moreover, we used the GNU \CC Compiler 5.5.0 with the flag \texttt{--O2}. 

\xhdr{GNNs.} 
For comparison with kernel methods, we used \textsc{Gin-$\varepsilon$}~\cite{Xu+2018b} and \textsc{Gin-$\varepsilon$-JK} with jumping knowledge networks as neural baselines~\cite{Xu+2018}. For data with (continuous) edge features, we used a $2$-layer MLP to map them to the same number of components as the node features and combined them using summation (\textsc{Gine-$\varepsilon$} and \textsc{Gine-$\varepsilon$}-JK). We used mean pooling to pool the learned node embeddings to a graph embedding and used a $2$-layer MLP for the final classification, using a dropout layer with $p = 0.5$ after the first layer of the MLP. For the smaller datasets of~\cref{t2l}, we optimized the number of hidden units from $
\{ 32, 64, 128\}$, the number of layers from $ \{1, 2, 3, 4, 5\}$ using the validation set. For the mid-scale datasets, due to computation time constraints, we set the number of hidden units to $64$ and the number of layers to $3$. Moreover, for both, we use a learning rate decay of $0.5$ with a patience parameter of $5$, a starting learning rate of $0.01$ and a minimum of $10^{-6}$, and a maximum epoch number of $200$. For both methods, we used the evaluation procedure described in~\cref{evalm} to optimize hyperparameters and compute accuracies. See the appendix for details on the hyperparameter and evaluation protocols used for the larger molecular regression tasks (\textsc{Zinc}, and \textsc{Alchemy}, \textsc{QM9}).

\xhdr{Results and discussion.}
\cref{t2l,t2ll,neural_short_tt} summarize the results. On the small-scale datasets, see \cref{t2l}, the \textsc{WL-OA} performs best overall. However, it does not scale to large datasets, \cref{t2ll}, as it relies on Gram matrix computation. Here, the $1$-WL performs well on all datasets, excluding \textsc{github\_stargazers}, where the neural baselines perform best overall.\footnote{For the neural baselines unlike the kernel baselines, we used one-hot degree features for datasets that do not provide node labels.}
Our results show that despite the extensive research on GNNs in recent years, classical graph kernels in combination with SVMs are still highly competitive in graph classification tasks.
\begin{table}[]\centering	
	\caption{Mean MAE (mean std. MAE, logMAE) on large-scale (multi-target) molecular regression tasks.\label{neural_short_tt}}
	\resizebox{1.0\columnwidth}{!}{\renewcommand{\arraystretch}{1.05}
		\begin{tabular}{lccc}	\toprule 
			\multirow{3}{*}{\vspace*{4pt}\textbf{Method}}&\multicolumn{3}{c}{\textbf{Dataset}}\\\cmidrule{2-4}
			& {\textsc{Zinc}}     & {\textsc{alchemy}}   & {\textsc{Qm9}}   \\	\toprule
			\textsc{Gine-$\varepsilon$}    & 0.084                                                                                                                                                                                                                                     \scriptsize $\pm 0.004$   &                                                                                                                                                                                                                                0.103 {\scriptsize $\pm 0.001$} -2.956 {\scriptsize $\pm 0.029$} &  0.081 {\scriptsize $\pm 0.003$} -3.400 {\scriptsize $\pm  0.094 $}   \\
			\textsc{MPNN}    & ---                                                                                                                                                                                                                                    & --- & 0.034 {\scriptsize $\pm 0.001$} {-4.156}{\scriptsize $\pm 0.030$}  \\		
			\bottomrule
	\end{tabular}}
\end{table}		
On the large-scale molecular learning tasks, see \cref{neural_short_tt}, it becomes apparent that specialized architectures such as \textsc{MPNN} result in significant gains over the generic \textsc{Gine-$\varepsilon$} baseline.

\section{Conclusion}

We gave an overview of the \textsc{TUDataset} collection, and reported on the results of an experimental study comparing graph kernels and GNNs on a subset of the data. We believe that our dataset collection will spark further progress in graph representation learning, and that our unified evaluation procedures will improve the comparability of results. We are looking forward to adding more datasets and are excited about contributions from the community, researchers, and practitioners from other areas. Future work includes a more extensive comparision of kernel and neural approaches on large-scale molecular regression tasks with continuous node and edge features.

\section{Acknowledgement}

We thank everybody who provided datasets for the  \textsc{TUDataset} collection.

This work has been partially funded by the Deutsche Forschungsgemeinschaft (DFG) within 
the Collaborative Research Center SFB 876 ``Providing Information by 
Resource-Constrained Data Analysis'', project A6 ``Resource-efficient Graph Mining''.
Nils Kriege has been supported by the Vienna Science and Technology Fund (WWTF) 
through project VRG19-009.

\bibliography{bibliography}
\bibliographystyle{icml2020}

\appendix

\onecolumn

\section{Evaluation examples}\label{evalcode}

See \url{www.graphlearning.io} for further documentation.

\xhdr{Kernelized SVM for graph kernels based on Gram matrices}
\begin{lstlisting}[language=Python]
import auxiliarymethods.auxiliary_methods as aux
import auxiliarymethods.datasets as dp
import kernel_baselines as kb
from auxiliarymethods.kernel_evaluation import kernel_svm_evaluation

# Download dataset.
classes = dp.get_dataset("ENZYMES")
use_labels, use_edge_labels = True, False

all_matrices = []
# Compute 1-WL kernel for 1 to 5 iterations.
for i in range(1, 6):
    # Use node labels and no edge labels.
    gm = kb.compute_wl_1_dense("ENZYMES", i, use_labels, use_edge_labels)
    # Apply cosine normalization.
    gm = aux.normalize_gram_matrix(gm)
    all_matrices.append(gm)

# Perform 10 repetitions of 10-CV using LIBSVM.
print(kernel_svm_evaluation(all_matrices, classes,
                            num_repetitions=10, all_std=True))

\end{lstlisting}

\xhdr{Linear SVM for graph kernels based on sparse feature maps}
\begin{lstlisting}[language=Python]
import auxiliarymethods.auxiliary_methods as aux
import auxiliarymethods.datasets as dp
import kernel_baselines as kb
from auxiliarymethods.kernel_evaluation import linear_svm_evaluation

# Download dataset.
classes = dp.get_dataset("MOLT-4")
use_labels, use_edge_labels = True, True

all_matrices = []
# Compute 1-WL kernel for 1 to 5 iterations.
for i in range(1, 6):
    # Use node labels and edge labels.
    gm = kb.compute_wl_1_sparse(dataset, i, use_labels, use_edge_labels)
    # Apply \ell_2 normalization.
    gm_n = aux.normalize_feature_vector(gm)
    all_matrices.append(gm_n)

# Perform 10 repetitions of 10-CV using LIBINEAR.
print(linear_svm_evaluation(all_matrices, classes,
                            num_repetitions=10, all_std=True))
\end{lstlisting}

\xhdr{GNN evaluation}
\begin{lstlisting}[language=Python]
import auxiliarymethods.datasets as dp
from auxiliarymethods.gnn_evaluation import gnn_evaluation
from gnn_baselines.gnn_architectures import GIN

dataset = "PROTEINS"
use_labels = True

# Download dataset.
dp.get_dataset(dataset)

# Optimize the number of layers ({1,2,3,4,5} and
# the number of hidden features ({32,64,128}),
# set the maximum nummber of epochs to 200,
# batch size to 64,
# starting learning rate to 0.01, and
# number of repetitions for 10-CV to 10.
print(gnn_evaluation(GIN, dataset, [1, 2, 3, 4, 5], [32, 64, 128], max_num_epochs=200,
                     batch_size=64, start_lr=0.01, num_repetitions=10, all_std=True))
\end{lstlisting}

\xhdr{Loading graphs in NetworkX format}
\begin{lstlisting}[language=Python]
import auxiliarymethods.datasets as dp
from auxiliarymethods.reader import tud_to_networkx

dataset = "PROTEINS"

# Download dataset.
dp.get_dataset(dataset)
# Output dataset as a list of graphs.
graph_db = tud_to_networkx(dataset)

\end{lstlisting}

\section{Experimental protocol and hyperparameters for \textsc{Zinc}, \textsc{Alchemy}, \textsc{QM9}}\label{eval}
For the larger molecular regression tasks, \textsc{Zinc} and \textsc{Alchemy},\footnote{Note that the full dataset is different from the contest dataset, e.g., it does not provide normalized targets, see \url{https://alchemy.tencent.com/}.} we closely followed the hyperparameters found in~\cite{Dwi+2020} and~\cite{Che+2020}, respectively, for the \textsc{Gine-$\varepsilon$} layers. That is, for \textsc{Zinc}, we used four \textsc{Gine-$\varepsilon$} layers with a hidden dimension of 256 followed by batch norm and a $4$-layer MLP for the joint regression of the twelve targets, after applying mean pooling. For \textsc{Alchemy} and \textsc{QM9}, we used six layers with 64 (hidden) node features and a set2seq layer~\cite{Vin+2016} for graph-level pooling, followed by a $2$-layer MLP for the joint regression of the twelve targets.

For \textsc{Zinc}, we used the given train, validation split, test split, and report the MAE over the test set. For the \textsc{Alchemy} and \textsc{QM9} datasets, we uniformly and at random sampled 80\% of the graphs for training, and 10\% for validation and testing, respectively. Moreover, following~\cite{Che+2020,Gil+2017}, we normalized the targets of the training split to zero mean and unit variance. We used a single model to predict all targets. Following~\cite{Kli+2020}, we report mean standardized MAE and mean standardized logMAE. We repeated each experiment five times (with different random splits in case of \textsc{Alchemy} and \textsc{QM9}) and report average scores and standard deviations.  Moreover, we use a learning rate decay of $0.5$ with a patience parameter of $5$, and a starting learning rate of $0.001$ with a minimum of $10^{-6}$.

For the \textsc{QM9} dataset, we additionally used the \textsc{MPNN}~\cite{Gil+2017} architecture as a baseline, closely following the setup of~\cite{Gil+2017}. For the \textsc{Gine-$\varepsilon$}  and the \textsc{MPNN} architecture, following~\citeauthor{Gil+2017}~\cite{Gil+2017}, we used a complete graph, computed pairwise $\ell_2$ distances based on the 3D-coordinates, and concatenated them to the edge features. We note here that our intent is not to beat the state-of-the-art, physical knowledge-incorporating architectures, e.g., \textsc{DimeNet}~\cite{Kli+2020} or \textsc{Cormorant}~\cite{And+2019}, but to solely provide baseline scores. 

All neural experiments were conducted on a workstation with four Nvidia Tesla V100 GPU cards with 32GB of GPU memory running Oracle Linux Server 7.7.
\newpage
\section{Dataset statistics}
\begin{table}[h]
	\begin{center}
		\caption{Dataset statistics and properties, $^\dagger$---Continuous vertex labels following~\cite{Gil+2017}, the last three components encode 3D coordinates.}
		\resizebox{1.0\textwidth}{!}{ 	\renewcommand{\arraystretch}{1.05}
			\begin{tabular}{@{}lcccccc@{}}\toprule
				\multirow{3}{*}{\vspace*{4pt}\textbf{Dataset}}&\multicolumn{6}{c}{\textbf{Properties}}\\
				\cmidrule{2-7}
				& Number of  graphs & Number of classes/targets & $\varnothing$ Number of vertices & $\varnothing$ Number of edges & Vertex labels & Edge labels \\ \midrule
				$\textsc{Enzymes}$       & 600               & 6                 & 32.6                             & 62.1                          & \cmark  & \xmark           \\
				$\textsc{IMDB-Binary}$   & 1\,000              & 2                 & 19.8                             & 96.5                          & \xmark   & \xmark          \\
				$\textsc{IMDB-Multi}$         & 1\,500               & 3                & 13.0                             & 65.9                          & \xmark    & \xmark         \\
				$\textsc{NCI1}$          & 4\,110              & 2                 & 29.9                             & 32.3                          & \cmark   & \xmark          \\
				$\textsc{NCI109}$        & 4\,127              & 2                 & 29.7                             & 32.1                          & \cmark    & \xmark         \\
				$\textsc{PTC\_FM}$       & 349               & 2                 & 14.1                             & 14.5                          & \cmark    & \xmark         \\
				$\textsc{Proteins}$      & 1\,113              & 2                 & 39.1                             & 72.8                          & \cmark    & \xmark         \\
				$\textsc{Reddit-Binary}$ & 2\,000              & 2                 & 429.6                            & 497.8                         & \xmark     & \xmark        \\ \midrule
							
				$\textsc{MCF-7}$       & 27\,770  & 2	&	26.4	&	 28.5               & \cmark  & \cmark           \\
					
				$\textsc{Molt-7}$       & 39\,765  &	2	&	26.1	&	     	28.1          & \cmark  & \cmark           \\
				$\textsc{Yeast}$       &   79\,601&	2&	21.5&	22.8                & \cmark  & \cmark           \\
				$\textsc{github\_star}$       &  12\,725 & 2	& 	113.8		&	234.6               & \xmark  & \xmark           \\
				$\textsc{reddit\_threads}$   &	203\,088	  & 2  &	23.9 &	25.0  		               & \xmark  & \xmark           \\
				\midrule        
				$\textsc{Zinc}$       &  249\,456 &12	&23.1 &	24.9     & \cmark  & \cmark           \\
				$\textsc{Alchemy}$       & 202\,579 &12	& 10.1 &	10.4 	     & \cmark  & \cmark           \\
				$\textsc{QM9}$       &129\,433  &12	& 18.0 &	18.6     & \cmark (13+3D)$^\dagger$  & \cmark (4)          \\
				\bottomrule
		\end{tabular}}
		\label{ds}
	\end{center}
\end{table}

\end{document}